\documentclass[11pt]{article}
\usepackage{coling2020}
\usepackage{latexsym}
\usepackage{url}

\usepackage[italian,french,spanish,english]{babel}
\usepackage[T1]{fontenc}
\usepackage[utf8x]{inputenc}
\usepackage{textcomp}
\usepackage{times}
\usepackage[colorlinks=true,citecolor=blue,urlcolor=blue,linkcolor=blue]{hyperref}
\usepackage{graphicx}
\usepackage{multirow,multicol}
\usepackage{multirow}
\usepackage[table,xcdraw]{xcolor}
\definecolor{darkgreen}{HTML}{006400}
\usepackage{xstring}
\usepackage{tikz-dependency}
\usepackage{soul}
\usepackage{float,lscape}
\usepackage{multicol}
\usepackage{footnote}
\usepackage{booktabs}
\usepackage{tablefootnote}

\usepackage{amssymb}
\usepackage{pifont}
\newcommand{\cmark}{\ding{51}}

\title{Multilingual Irony Detection with Dependency Syntax and Neural Models}

\author{Alessandra Teresa Cignarella$^{1,2}$,
  Valerio Basile$^{1}$,
  Manuela Sanguinetti$^{3}$,
  Cristina Bosco$^{1}$,\\
  \textbf{Paolo Rosso}$^{2}$
  \textbf{and Farah Benamara}$^{4}$\\
  1. Dipartimento di Informatica, Università degli Studi di Torino, Italy \\
  2. PRHLT Research Center, Universitat Politècnica de València, Spain \\
  3. Dipartimento di Matematica e Informatica, Università degli Studi di Cagliari, Italy\\
  4. IRIT-CNRS, Université de Toulouse, France \\
  \texttt{\small \{cigna,basile,bosco\}@di.unito.it}, \texttt{\small manuela.sanguinetti@unica.it},\\
  \texttt{\small prosso@dsic.upv.es}, \texttt{\small farah.benamara@irit.fr}}

\date{}

\begin{document}
\maketitle
\begin{abstract}
This paper presents an in-depth investigation of the effectiveness of dependency-based syntactic features on the irony detection task in a multilingual perspective (English, Spanish, French and Italian). It focuses on the contribution from syntactic knowledge, exploiting linguistic resources where syntax is annotated according to the \textit{Universal Dependencies} scheme.
Three distinct experimental settings are provided. In the first, a variety of syntactic dependency-based features combined with classical machine learning classifiers are explored. In the second scenario, two well-known types of word embeddings are trained on parsed data and tested against gold standard datasets. 
In the third setting, dependency-based syntactic features are combined into the Multilingual BERT architecture. The results suggest that fine-grained dependency-based syntactic information is informative for the detection of irony.
\end{abstract}

{\let\thefootnote\relax\footnotetext{Copyright \textcopyright\ 2020 for this paper by its authors. Use permitted under Creative Commons License Attribution 4.0 International (CC BY 4.0).}}

\section{Introduction}\label{sec:intro}
Irony is a pragmatic phenomenon that can be defined as a sort of incongruity between the literal meaning of an utterance and its intended meaning \cite{grice1975logic,sperber1981irony,attardo2000irony}. In text processing, the challenging nature of irony is mainly due to the inherent features of the phenomenon itself, that may assume a large variety of merged forms and facets. Several semantic and syntactic devices can be used to express irony (e.g., analogy, oxymoron, paradox, euphemism and rhetorical questions), and a variety of different linguistic elements can cause the incongruity, determine the clash and play the role of irony triggers within a text.

Most computational research in irony detection is applied on social media texts and focuses primarily on content-based processing of the linguistic information using semantic or pragmatic devices, neglecting syntax.
For example, \newcite{joshi2015harnessing} 
propose a method for sarcasm detection in tweets as the task of detecting incongruity between an observed and an expected word, while \newcite{karoui2015towards} rely on external common knowledge to the utterance shared by 
author and 
reader.
However, in addition to these pragmatic devices, 
some hints about the role of syntax can be found in linguistic literature where the deviation from syntactic norms has been reported as a possible trigger of the phenomenon.
{For instance, \newcite{michaelis2015whatisthis} explore the usage of \textit{split interrogatives} and of \textit{topic-comment utterances} in English, arguing they both have a double function (they express two distinct communicative acts) and that one of those is precisely that of conveying an ironic meaning (e.g. ``Who are you, Willy Wonka?'').\footnote{\newcite{michaelis2015whatisthis} use the term ``sarcasm'', but for the purposes of the present paper we can gloss over the subtle differences between the two phenomena and consider \textit{irony} as the most general term that also includes sarcasm beneath it.}}
The interpretation of an ironic utterance involves not only the knowledge of the usage of the language, inventory and rules including grammar, lexis, syntax, structure and style, but also the perception of the violation of these rules. Irony tactics handle the syntagmatic patterns of discourse, the way in which items precede, follow and are related to each other. As \newcite{chakhachiro2018translating-irony} shows referring to Arabic, syntactic norms, which normally see adverbs placed after verbal phrases, can be violated to trigger irony. However, when the commonly used ironic device of topic shift occurs, lexical and syntactic signals are always present to mark it. 
The fact that the expression of opinion or the production of irony are sensitive to syntactic variations occurring in texts has been moreover confirmed by \newcite{mahler2017breaking}, where artificial syntactic manipulations are applied, such as those based on negations and adverbs, with the purpose of fooling sentiment analysis systems.

Some research already explored different kinds of syntactic features and their interaction in several NLP tasks, showing their effectiveness. For example, \newcite{sidorov2012syntactic} exploited 
syntactic dependency-based n-grams 
for general-purpose classification tasks, 
\newcite{Socher2013} investigated 
sentiment and syntax 
with the development of a sentiment treebank, 
and \newcite{kanayama-iwamoto-2020-universal} showed 
a pipeline method that makes the most of syntactic structures based on Universal Dependencies, achieving high precision in sentiment detection for 17 languages.
Morphology and syntax have also been proved useful in a number of other tasks, such as rumor detection \cite{ghanem2019upv}, authorship attribution \cite{posadas2014complete,sidorov2014syntactic} and
humor detection \cite{liu2018exploitingsyntacticstructuresforhumor}. To the best of our knowledge, very few studies use syntactic information specifically for irony detection. Among them we cite  \newcite{cignarella2019atc}, who employed a SVC combined with shallow features based on morphology and dependency syntax outperforming strong baselines in the IroSvA 2019 irony detection shared task in Spanish variants \cite{ortega2019overview}.

This paper aims to go one step further focusing for the first time on the development of syntax-aware irony detection systems in a multilingual perspective (English, Spanish, French and Italian), providing an in-depth investigation of the impact of different sources of syntactic information when used on top of several machine learning models, including Recurrent Neural Networks and Transformers.
It is important to note that our aim is not
to outperform the current state of the art on monolingual irony detection, but to investigate whether irony detection based on \textit{syntax alone} can achieve comparable results with existing systems when evaluated in \textit{multiple languages}. We believe this is an important first step before moving to complex scenarios where both syntax and pragmatic knowledge are incorporated into deep learning models which could lead to explicitly model the syntax-pragmatic interface of irony. 
To this end, we propose to address the following research questions:

\begin{itemize}
    \item \textbf{(RQ-1)} \emph{Can morphological and syntactic knowledge be helpful in addressing the task of irony detection?} - We exploit linguistic resources syntactically annotated according to a well-known dependency-based scheme (\textit{Universal Dependencies}\footnote{\url{https://universaldependencies.org/}.} UD).
    The versatility of the UD format allows us to apply our approaches for the detection of irony  independently of a target language. Furthermore, we experimented with UD-based word embeddings, and to the best of our knowledge, neither UD annotations nor syntax-based pre-trained embeddings have previously been used for irony detection.
    
    \item \textbf{(RQ-2)} \emph{To what extent do \emph{ad-hoc} resources in UD format (treebanks) improve irony detection performances?} - We propose three  distinct experimental settings.  Firstly, a variety of syntactic dependency-based features combined with classical machine learning classifiers are explored, with the aim of finding the most informative set of features for detecting the presence of irony.  In the second scenario two well-known word-embedding models are tested against gold standard datasets.  Finally, in the third setting, dependency-based syntactic features are combined into the Multilingual BERT architecture.
    Our results show that our models outperform syntax-agnostic ones, which is an important first step towards syntactically-informed irony detection.
    
    \item \textbf{(RQ-3)} \emph{Are results obtained using syntactic features stable across different languages?} - We experiment with datasets made available from previous shared tasks on irony detection in four languages: French (DEFT 2017 \cite{DEFT2017}), English (SemEval-2018 Task 3 \cite{semeval2018ironytask}), Spanish (IroSvA \cite{ortega2019overview}) and Italian (IronITA \cite{cignarella2018overview}). 
    The obtained results overcome the competitive baselines and are also favorably comparable with the best results reported in shared tasks on irony for the four studied languages.
    On the other hand, this study takes the opportunity to enrich such existing datasets with morphological and syntactic information, as encoded in UD. 
\end{itemize}


\noindent The paper is organized as follows. In the next section, related work regarding irony detection is surveyed,
while Section \ref{sec:data-experimental-setting} describes the data and experimental settings we followed. In Section \ref{sec:methodology}, we provide a description of the experiments we performed, and, in the last section, we discuss the experiments and provide some final remarks through an error analysis. We conclude the paper with some insights about the impact of morpho-syntax information and suggest possible directions for future work.

\section{Related work}\label{sec:related-work}
The identification of irony and the description of pragmatic and linguistic devices that trigger it has always been a controversial topic \cite{grice1975logic,sperber1981irony,utsumi1996unified}, a challenge for both humans and automatic tools. This motivates the interest for irony of the NLP community, the organization of tasks 
regarding its processing and the subsequent creation of benchmarks.



In the last decade several shared tasks have been organized in order to advance with the techniques regarding the automatic processing of figurative language \cite{ghosh2015semeval}, and more recently irony and sarcasm detection have been gaining greater attention. Being irony a pragmatic device inherently related to culture and language, the task captured the attention of various researchers, dealing with different languages and mostly focusing on social media texts.
The presence of irony for Italian has been investigated -among other things- in the SENTIPOLC task of the EVALITA 2014 \cite{basile2014overview} and 2016 \cite{sentipolc2016} and at IronITA \cite{cignarella2018overview}. It was addressed for French within the context of DEFT 2017 \cite{DEFT2017}, for English at Semeval 2018 (\textit{Task 3: Irony Detection in English Tweets})  \cite{semeval2018ironytask} as well as for Spanish at the IroSVA \cite{gonzalez2019elirfirosva}. As a side effect, several annotated corpora were released, among which also those that allowed us the composition of the experimental settings provided in this paper (see \S\ \ref{sec:methodology}). 

Similar to other NLP tasks, neural approaches in irony detection have been widely used, and proved successful even in the mentioned shared tasks. For instance, the best-scoring team for Task 3 at Semeval 2018 used a densely connected LSTM based on pre-trained word embeddings, sentiment and PoS tag features \cite{wu2018thu_ngnSemEval}; they built a multi-task model to predict the missing irony hashtag, whether a tweet is ironic or not and the fine-grained type of irony. A similar multi-task learning approach was adopted in \newcite{cimino2018multitasklearning} at IronITA 2018, which introduced a 2-layer BiLSTM that exploits additional information, such as automatically-generated sentiment polarity lexica, word-embedding lexica and PoS tags. The irony detection system proposed in \newcite{gonzalez2019elirfirosva} at IroSVA 2019 was based on the Transformer Encoder model. At DEFT 2017 the winning team of the shared task for French \cite{rouvier2017liadeft} proposed an approach based on varying the size of the layers of a
CNN combined with three different sentiment-based word embeddings.
Finally, a multilingual scenario is also proposed in \newcite{ghanem2020irony}, that describe both feature-based models (with Random Forest showing the best results) and a CNN architecture with multilingual embeddings, and test their approach on Arabic, English and French (for the latter, the DEFT 2017 dataset was used, as in the present work).
As mentioned in Section \ref{sec:intro}, to the best of our knowledge, very few studies use syntactic information specifically for irony detection.

While in this work as well we tested our models in a multilingual setting,
the novelty of the approach provided in this paper consists in using features which represent syntactic knowledge that can be extracted from text by applying morphological and syntactic analysis.
Universal Dependencies 
played a key role in this respect: in recent years they have become a popular and widely-used annotation format \cite{UD2020}, providing a set of broadly attested universal grammatical relations, in which existing dependency schemes for different languages can be mapped.
The proliferation of UD treebanks in a short time span highlights the need for a consistent representation of linguistic phenomena, not only for typical contrastive studies, but also
when using such resources for downstream applications.

\section{Data and Experimental Setting}\label{sec:data-experimental-setting}
In order to address the research questions, we follow a binary classification irony detection task, testing an automatic system on four different languages: English, Spanish, French and Italian. 
This will allow us to investigate whether syntactic structures, independently of the target language, are informative to understand whether a message is ironic or not. 

In the multilingual experimental setting we took advantage of four datasets that have been made available during the last few years within evaluation campaigns, and all featured
by a binary annotation for irony. In Table \ref{tab:datasets-A}, for each dataset, we report the target language, the name of the shared task in which it was released and its reference, the number of tweets for each class (\textit{ironic} vs. \textit{not ironic}) and the total number of instances, for both training set and test set.
The aim of our task is, thus, to determine whether a given tweet is \textit{ironic} or \textit{not ironic}.

\begin{table}[!ht]
\footnotesize 
\centering
\begin{tabular}{llrrr|rrr}
\textbf{language} & \textbf{dataset} & \multicolumn{3}{c}{\textbf{train}} & \multicolumn{3}{c}{\textbf{test}} \\
\toprule
&  & {ironic} & {not} & {total} & {ironic} & {not} & {total} \\
\midrule
English & SemEval-2018 Task 3 \cite{semeval2018ironytask}
& 1,923 & 1,911 & 3,834 & 311 & 473 & 784 \\
Spanish & IroSvA 2019 \cite{ortega2019overview}
& 1,600 & 5,600 & 7,200 & 599 & 1,201 & 1,800   \\
French  & DEFT 2017 \cite{DEFT2017}
& 1,947 & 3,906 & 5,853 & 488 & 976 & 1,464 \\
Italian & IronITA 2018 \cite{cignarella2018overview}
& 2,023 & 1,954 & 3,977 & 435 & 437 & 872 \\ 
\bottomrule
\end{tabular}
\caption{Benchmark datasets used for irony detection (binary task).}\label{tab:datasets-A}
\end{table}

\noindent  In Table \ref{tab:examples-iro-not}, we show a pair of ironic and non-ironic tweets for each language, with examples retrieved from the datasets used in the present study.

\begin{table}[!ht]
\centering
\footnotesize
\begin{tabular}{llc}
\textbf{language} & \textbf{tweet} & \textbf{irony} \\

\toprule

\multirow{2}{*}{English}
& Just great when you're mobile bill arrives by text & ironic \\
\cmidrule{2-3}
& i feel like whole life is about waiting waiting and waiting & not \\

\midrule

\multirow{4}{*}{Spanish} & \#Bienvenidos a \#Carmenolandia. Ciudad del \#Atasco perpetuo. \#SemáforosA5 & \multirow{2}{*}{ironic} \\
& \textit{\#Welcome to \#Carmenoland. City of the infinite rush hour. \#TrafficLightsA5} & \\
\cmidrule{2-3}
& La fotografía que ilustra los \#SemáforosA5 \url{https://t.co/qKLXgr6jJF} & \multirow{2}{*}{not} \\
& \textit{The photograph that shows the traffic lights A \url{https://t.co/qKLXgr6jJF}} & \\

\midrule

\multirow{4}{*}{French}
& Hollande et Sarkozy au second tours, on s'attendez pas a ça & \multirow{2}{*}{ironic} \\
& \textit{Hollande and Sarkozy in the second round election, we didn't expect it} & \\
\cmidrule{2-3}
& Le football c'est pas cruel, c'est juste la loi du plus fort & \multirow{2}{*}{not} \\
& \textit{Football is not cruel, it's just the law of the strongest} & \\

\midrule   

\multirow{4}{*}{Italian}
& Nuovo governo monti: dal bunga bunga al banca banca... & \multirow{2}{*}{ironic} \\
& \textit{New Monti's Governmet: from bunga bunga to bank bank...} & \\
\cmidrule{2-3}
& REUTERS ANALISI Crisi, di corsa verso il governo Monti & \multirow{2}{*}{not} \\
& \textit{REUTERS ANALYSIS Crisis, running towards Monti's Government}  &  \\
\bottomrule
\end{tabular}
\caption{Examples from the datasets.}
\label{tab:examples-iro-not}
\end{table}

\begin{figure}[!ht]
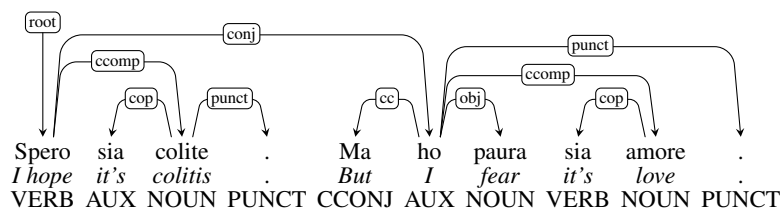

\centering
\begin{dependency}
\small
\begin{deptext}
Spero	\& sia	\& {colite}	\& .	\& Ma	\& ho	\& paura	\& sia	\& amore	\& .	\\
\textit{I hope}	\& \textit{it's}	\& \textit{colitis}	\& .	\& \textit{But}	\& \textit{I}	\& \textit{fear}	\& \textit{it's}	\& \textit{love}	\& .	\\
VERB \& AUX \& NOUN \& PUNCT \& CCONJ \& AUX \& NOUN \& VERB \& NOUN \& PUNCT\\
\end{deptext}
\deproot[edge unit distance=3ex]{1}{root}
\depedge{3}{2}{cop}
\depedge{1}{3}{ccomp}
\depedge{3}{4}{punct}
\depedge{6}{5}{cc}
\depedge[edge unit distance=1.9ex]{1}{6}{conj}
\depedge{6}{7}{obj}
\depedge{9}{8}{cop}
\depedge[edge unit distance=1.9ex]{6}{9}{ccomp}
\depedge[edge unit distance=2.1ex]{6}{10}{punct}
\end{dependency}
\caption{The dependency tree in UD format.}\label{fig:UD-tree}
\end{figure}

\noindent Provided that the availability of morphological and syntactic knowledge is crucial for performing the experiments described in the rest of the paper, we needed to obtain a representation of all the datasets in UD format.
With the exception of \textsc{twittirò} \cite{cignarella2019presenting}, a subset of the IronITA dataset already available as a UD treebank (see the current release in the UD official repository\footnote{\url{https://github.com/UniversalDependencies/UD_Italian-TWITTIRO}.}), 
we obtained the dependency-based annotation for the other corpora by applying the \textit{UDPipe}\footnote{\url{http://ufal.mff.cuni.cz/udpipe}.} pipeline (for tokenization, PoS-tagging and parsing). 
Figure \ref{fig:UD-tree} provides an example drawn from \textsc{twittirò}.

\noindent Considering that all the datasets used in this work consist of Twitter data, whenever possible, we used resources where this genre, or at least user-generated content of some kind, was included as training data for parsing.
More precisely, the model for English has been trained on the EWT treebank \cite{silveira2014gold}, that for Spanish on both GSD-Spanish \cite{mcdonald2013universal} and ANCORA corpora \cite{taule2008ancora}. The model for Italian -- for the remaining part of IronITA\footnote{Approximately 1,400 out 4,849 tweets from IronITA are also part of the \textsc{twittirò} corpus, already available in UD.} -- was trained on POSTWITA \cite{sanguinetti2018postwita} and ISDT treebanks \cite{simi2014lessismore}, while that for French on the GSD-French corpus \cite{mcdonald2013universal}.
 We are aware that there actually exists a Twitter treebank for English, i.e. TWEEBANK V2 \cite{liu2018parsingtweets}, but it is not fully compliant with the UD format specifications (e.g. it violates the single root constraint posed in UD format). We thus opted for the EWT in order to preserve annotation consistency among resources.

The different amount of data used for training UDPipe, the variety of text genres among the datasets, and the fact that the UD annotation of only one dataset has undergone a manual correction -- i.e. \textsc{twittirò}, also mentioned above -- can make the quality of the UD data used in this study not entirely homogeneous. 
In spite of such disparity in the annotation reliability, and bearing in mind that a higher accuracy in this regard can be crucial, we considered the output provided by the parser for all the languages reasonably satisfactory for the purposes of our study.

\section{Methodology}\label{sec:methodology}
The main aim of the experiments presented in this section consists in evaluating the contribution to irony detection made by the linguistic information provided in the datasets described above. 
%
The task we propose is, therefore, a straightforward binary classification task on irony detection, that is, the task for which literature offers baselines and fair-sized annotated datasets for a variety of languages.


\noindent For addressing the task, we performed a set of experiments where several models were implemented exploiting classical machine learning algorithms, deep learning architectures and state-of-the-art language models implemented with the Python libraries \emph{scikit-learn}\footnote{\url{https://scikit-learn.org}.} and \emph{keras}\footnote{\url{https://keras.io/}.}. We tested different sets of pre-trained word embeddings to initialize the neural models, namely \emph{fastText}\footnote{\url{https://fasttext.cc/}.} and a dependency-based \emph{word2vec} proposed by \newcite{levy2014dependencywordembeddings} (\emph{word2vecf}). The latter was trained on the concatenation of all the treebanks available in the UD repository for each considered language.

In order to combine these methods with syntactic features inspired by \newcite{sidorov2014syntactic}, we used data where not only a binary annotation for irony is applied, but also a morphological and syntactic analysis is available (see \S\ \ref{sec:methodology}).

\subsection{Pre-processing and Features}\label{sec:features}
\noindent We stripped all the URLs and we normalized all characters to lowercase letters, as it is often done before the application of sentiment analysis tools. We investigated the use of novel features aimed at exploiting information conveyed by syntax, studying in particular the impact of the availability of training resources in UD format.

\begin{figure}[!h]
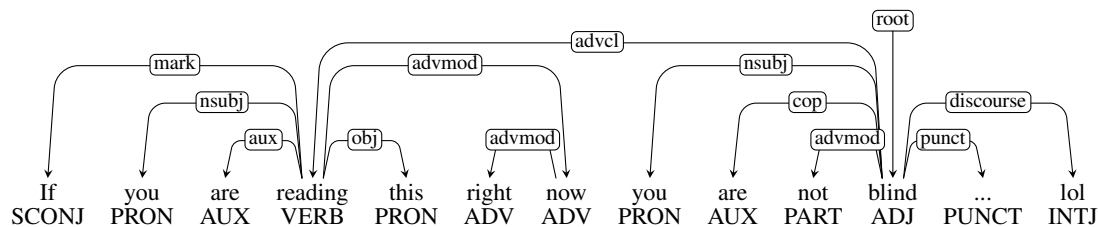

\centering
\small
\begin{dependency}
\begin{deptext}
If  \&[.2cm] you  \&[.2cm] are  \&[.2cm] reading  \&[.2cm] this  \&[.2cm] right  \&[.2cm] now  \&[.2cm] you  \&[.2cm] are  \&[.2cm] not  \&[.2cm] blind  \&[.2cm] ...  \&[.2cm] lol \\
SCONJ  \&[.2cm] PRON  \&[.2cm] AUX  \&[.2cm] VERB  \&[.2cm] PRON  \&[.2cm] ADV  \&[.2cm] ADV  \&[.2cm] PRON  \&[.2cm] AUX  \&[.2cm] PART  \&[.2cm] ADJ  \&[.2cm] PUNCT  \&[.2cm] INTJ \\ 
\end{deptext}
\depedge{4}{1}{\normalsize mark}
\depedge{4}{2}{\normalsize nsubj}
\depedge{4}{3}{\normalsize aux}
\depedge[edge unit distance=1.8ex]{11}{4}{\normalsize advcl}
\depedge{4}{5}{\normalsize obj}
\depedge{7}{6}{\normalsize advmod}
\depedge{4}{7}{\normalsize advmod}
\depedge{11}{9}{\normalsize cop}
\depedge{11}{8}{\normalsize nsubj}
\depedge{11}{10}{\normalsize advmod}
\depedge{11}{12}{\normalsize punct}
\depedge{11}{13}{\normalsize discourse}
\deproot[edge unit distance=4ex]{11}{\normalsize root}
\end{dependency}
\caption{Dependency-based syntactic tree of an English tweet.}\label{fig:syntax-tree}
\end{figure}

\noindent The description of features as well as the content of the vectors for the syntactic features we developed, referring to the tweet in Figure \ref{fig:syntax-tree}, are as follows:\\
\noindent $\bullet$ \textsf{ngrams}: we extracted unigrams, bigrams and trigrams of tokens; e.g. \textit{[If, you, are, reading, ..., If you, you are, are reading, ..., If you are, you are reading, are reading this, ...]};\\
$\bullet$ \textsf{chargrams}: we considered the sequence of char-grams in a range from 2 to 5 characters; eg \textit{[If, fy, yo, ou, ..., Ifyou, fyoua, youar, ouare, uarer, ...]};\\
$\bullet$ \textsf{deprelneg}: we considered the presence of negation in the text, relying on the morpho-syntactic cues present in the UD format. When a negation was present, we appended the corresponding dependency relation in the feature vector. For instance, in Figure \ref{fig:syntax-tree}, we spot a negation in \textit{[... are \textbf{not} blind ...]}, the dependency relation of ``not'' is \texttt{advmod}, therefore we append it in the feature vector; \\
$\bullet$ \textsf{deprel}: we built a bag of words of 5-grams, 6-grams and 7-grams of dependency relations as occurring in the linear order of the sentence from left to right; e.g. \textit{[mark nsubj aux obj advmod, nsubj aux obj advmod advmod, ..., advmod advmod nsubj cop advmod root punct, advmod nsubj cop advmod root punct discourse]};\\
$\bullet$ \textsf{relationformVERB}: we create a feature vector with all the tuples of tokens that are connected with a dependency distance $DD = 1$, by starting from a verb and at the same time we blank the verb itself. For instance, in our example the first verb is ``\textit{reading}''  and some of the tuples of tokens connected through this verb are, e.g. \textit{[IfVERBthis, youVERBthis, areVERBthis, IfVERBnow, youVERBnow, ...]};\\ 
$\bullet$ \textsf{relationformNOUN}: we applied the same procedure of the feature above but considering nouns as starting points for collecting tuples;\\
$\bullet$ \textsf{relationformADJ}: in the same fashion of the two features above, we repeated the same procedure for adjectives too; \\
$\bullet$ \textsf{Sidorovbigramsform}: we created a bag of word-forms (tokens), considering the bi-grams that can be collected following the syntactic tree structure (rather than the bi-grams that can be collected reading the sentence from left to right).\footnote{Please refer to \newcite{sidorov2013syntactic} and \newcite{sidorov2014should} for more details on this regard.} Such that: e.g. \textit{[blind reading, blind you, blind are, blind not, reading if, reading you,  ...]};\\
$\bullet$ \textsf{Sidorovbigramsupostag}: as the feature above, we created a bag of Part-of-Speech tags;\\
$\bullet$ \textsf{Sidorovbigramsdeprel}: as the two features above, we created a bag of words based on dependency relations (\textit{deprels}).\\

\subsection{Models}\label{subsec:models}
Having as primary goal the exploration of the features described in the previous section and as a case study testing their effectiveness in irony detection, we implemented a variety of models,
including the following\vspace{0.1cm}:

\noindent
{\textbf{Support Vector Machine (SVM) - } The choice of the kernel could be a challenging task, because the performance is dataset dependent.~In this case we opted for a linear kernel, which is usually recommended for text classification.\footnote{We also experimented with RBF kernel but it obtained worse performances due to sparse feature matrices.}}\\
\textbf{Logistic Regression (LR) - } We used the Logistic Regression classifier with the default parameters, exception made for the maximum number of iterations which we set to 5.\\
\textbf{Random Forest (RF) - } We used the Random Forest classifier with its default parameters.\\
{\textbf{Multi-Layer Perceptron (MLP) - } We used \textit{GridSearch}\footnote{\url{https://bit.ly/2NxU4xC}.} to adjust the hyper-parameters, thus obtaining: hidden layer size $ = 30$, the activation function for the hidden layer = $logistic$, the solver for weight optimization $ = lbfgs$. We later set the \emph{early stopping} to \texttt{True}, and an initial learning rate $ = .01$, and the size of minibatches $ = 5$.}

\noindent
\noindent \textbf{GRU - } We used a straightforward architecture and set the following hyper-parameters: epochs $= 10$, learning rate $= .0001$. The GRU is initialized with random weights and the word embeddings are learned during the training.\footnote{We experimented also with other RNN architectures but they were not conclusive.}\\
\noindent \textbf{GRU+fastText - } With the same hyper-parameters as above, we initialized the neural network with specific \emph{fastText} word embedding pre-trained models~\cite{joulin2016fasttext} for each language.\\
\noindent \textbf{GRU+dependency-based embeddings - } In the same way, and maintaining the same setting, we fed the neural network with the dependency-based \emph{word2vec} embeddings by \newcite{levy2014dependencywordembeddings}. These word embeddings are able to capture syntactic information during training, therefore producing embeddings that are more sensitive to functional similarity than traditional co-occurrence-based word embeddings like \emph{fastText}. We trained a different word embedding model for each language using the concatenation of available UD treebanks for that language.

\noindent
\noindent \textbf{Multilingual BERT - } Since a multilingual version of the BERT language model (henceforth M-BERT)\footnote{\url{https://github.com/google-research/bert/blob/master/multilingual.md}.} is also available for 104 languages including the four we take into account in this study, we also performed a set of experiments using it. We set the hyper-parameters such that the batch size $ = 8$, the initial learning rate $ = 1e-5$. We did not set any fixed number of epochs, but rather relied on the \emph{EarlyStopping} function, setting the value of patience $ = 5$.\\
\noindent \textbf{M-BERT+syntax - } With the same hyper-parameters as above, we concatenated the dependency-based features from \S\ \ref{sec:features} to those extracted from M-BERT and fed both of them into a LSTM neural network. \\
\noindent \textbf{M-BERT+best\_feats - } Maintaining the same setting as the two previous models, here we concatenated a smaller set of features, and not all of them. Namely, we selected only the \textit{best features} resulting from experiments with classical machine learning models, as we will explain in \S\ \ref{sec:results-classical-ML}. \\
\noindent \textbf{M-BERT+autoencoder - } Finally, we experimented by reducing the features with a technique
known as \textit{Autoencoder}~\cite{bengio2009learning} which is becoming more and more popular in NLP \cite{zabalza2016autoencoder}.

\section{Experiments and Results}
We performed different sets of experiments training the models on the training sets and evaluating them against gold test sets made available in the shared tasks referred to in Table \ref{tab:datasets-A}. We highlight three main scenarios:
\S\ \ref{sec:results-classical-ML} - experiments to select the best features; \S\ \ref{sec:results-deeplearning-and-WE} - experiments to measure the impact of dependency-based word embeddings; and finally \S\ \ref{sec:results-lang-models-autoencoders} - experiments conducted with the bidirectional encoder BERT in its multilingual variant infused with syntax knowledge.

\subsection{Selection of Best Features}\label{sec:results-classical-ML}
Firstly, in order to understand which of the features provided in \ref{sec:features} are relevant for the present task, we performed experiments with classical machine learning algorithms. We carried out an evaluation of four models (SVM, LR, RF and MLP) by combining them with all the possible permutations of the feature set and we evaluate them with respect to the macro-F1 score (the averaged score between the F1 of the ironic class and the F1 of the non-ironic one).\footnote{Supplementary material such as code, detailed features and exhaustive tables of results are accessible here: \url{https://github.com/AleT-Cig/DependencySyntax_DeepIrony}.}
This first step was crucial for establishing which single feature or group of combined features performs better for each of the involved languages, as summarized in Table \ref{tab:checkmarks}.

\begin{table}[!ht]
\resizebox{\textwidth}{!}{%
\begin{tabular}{lcccccccccccc}
\textbf{language} & \multicolumn{1}{c}{\textbf{\begin{tabular}[c]{@{}c@{}}macro\\ F1\end{tabular}}} & \textbf{model} & \multicolumn{1}{c}{\textbf{ngrams}} & \multicolumn{1}{c}{\textbf{chargrams}} & \multicolumn{1}{c}{\textbf{deprel}} & \multicolumn{1}{c}{\textbf{deprelneg}} & \multicolumn{1}{c}{\textbf{\begin{tabular}[c]{@{}c@{}}relation\\form\\ VERB\end{tabular}}} & \multicolumn{1}{c}{\textbf{\begin{tabular}[c]{@{}c@{}}relation\\form\\ NOUN\end{tabular}}} & \multicolumn{1}{c}{\textbf{\begin{tabular}[c]{@{}c@{}}relation\\form\\ ADJ\end{tabular}}} & \multicolumn{1}{c}{\textbf{\begin{tabular}[c]{@{}c@{}}Sidorov\\ bigrams\\form\end{tabular}}} & \multicolumn{1}{c}{\textbf{\begin{tabular}[c]{@{}c@{}}Sidorov\\ bigrams\\deprel\end{tabular}}} & \multicolumn{1}{c}{\textbf{\begin{tabular}[c]{@{}c@{}}Sidorov\\ bigrams\\upostag\end{tabular}}} \\

\toprule

{English} & .683 & RF &  & \cmark &  &  & \cmark &  & \cmark &  & \cmark & \cmark \\
{Spanish} & .539 & RF &  &  &  &  &  &  & \cmark & \cmark &  &  \\
{French} & .641 & LR &  & \cmark &  &  & \cmark & \cmark &  & \cmark &  & \cmark \\
{Italian} & .702 & RF &  & \cmark & \cmark &  & \cmark & \cmark &  & \cmark & \cmark &  \\
\bottomrule
\end{tabular}
}
\caption{Features exploited in the best runs with classical ML algorithms in each language scenario.}\label{tab:checkmarks}
\end{table}

\noindent From Table \ref{tab:checkmarks} it emerges how in all the configurations used for achieving the best score, at least one dependency-based syntactic feature was exploited and in particular those based on Sidorov's work, i.e. the last three columns of the table\footnote{Please refer to \newcite{sidorov2012syntactic} and \newcite{sidorov2013syntactic} for further details.}. This provides evidence for positively answering to our first research question, since those are the features where the real structure from root to branches of syntactic trees is encoded.

Moreover, Italian is the best scoring language (F1 = $.702$). This might provide some hints about our second research question, suggesting that the higher quality of the resource from where syntactic knowledge has been drawn might positively influence the performance.
As previously mentioned, the one for Italian is the only dataset used for this study previously submitted to a careful manual check, after the processing done with UDPipe. On the contrary, the datasets for the other three languages are obtained automatically without further manual revision.

Interestingly, in any configuration \emph{ngrams} and \emph{deprelneg} are used. Regarding the latter, we noticed that it was a weak feature, and very sparse, thus not able alone to capture a complex phenomenon as irony. Concerning \emph{ngrams}, although, the fact that they are not exploited in the best configurations might point towards the fact that to detect irony a lexical approach is not really sufficient.

Through this step we are able to consider these combinations of characteristics as ``\textit{best feature sets}'' for each language and to exploit them in further experiments (see \S\ \ref{sec:results-lang-models-autoencoders} below in this section).

\subsection{Measuring the Impact of Dependency-Based Word Embeddings}\label{sec:results-deeplearning-and-WE}
In the second set of experiments we explored, instead, the potential of two different kinds of word embeddings with regard to each language, namely \emph{fastText} and the dependency-based word embeddings version of \emph{word2vec}, which we trained on the four datasets taking advantage of the UD representation (\emph{dep-based we}).

\begin{table}[h]
\centering
\footnotesize
\begin{tabular}{llccccc}
\multirow{2}{*}{\textbf{language}} & \multicolumn{6}{c}{\textbf{GRU}} \\
\cmidrule{3-7}
& & &
{{\color{white}{---}} --- {\color{white}{---}}} &
&
\textbf{+fastText}
& \textbf{+dep-based we} \\
\toprule
English  & & & .648  & &  \textbf{.650}       &    .552        \\
\midrule
Spanish  & & & .494  & &  \textbf{.500}       &    .404        \\
\midrule
French   & & & .522  & &  \textbf{.567}       &    .447        \\
\midrule
Italian  & & & .649  & &  .652       &    \textbf{.659}        \\
\bottomrule
\end{tabular}
\caption{Results obtained combining a GRU architecture and word embeddings.}\label{tab:results-gru}
\end{table}

\noindent As depicted in Table \ref{tab:results-gru}, for each language, the best results in terms of macro F1 are obtained adopting either \emph{fastText} or the \emph{dependency-based word embeddings}.
Furthermore, we highlight that \emph{fastText} seems overall the best configuration in this set of experiments (English, Spanish and French), exception made for Italian in which the best result is obtained with the \emph{dependency-based word embeddings} configuration. 
The most significant insight that we get from this second set of experiments is that neural architectures (in this case GRUs) can benefit from the addition of syntactic features and from training on UD-based representations, {although they strictly depend on the quality of training data. In fact, from the results shown in Table \ref{tab:results-gru}, it can be seen that the performance of the dependency-based word embeddings is much lower than that of the randomly initialized word embeddings and the word embeddings pre-trained from fastText for most languages.
A positive result with respect to this is obtained in Italian, where the \emph{dependency-based word embeddings} configuration gives some hints about the impact that a good quality of the morpho-syntactic annotation might provide. The Italian dataset is, in fact, the only one amongst the four ones that has been manually checked and corrected and whose quality could be considered superior compared to the resources used for the other languages.
Furthermore, Italian is the only language configuration in which the \emph{dependency-based word embeddings} have been trained on UD-based text from the same textual genre (tweets).
To the best of our knowledge, \textit{dependency-based word embeddings} were not used before in the detection of irony; on the other hand, in a previous work, \newcite{macavaney2018deeperlook} expand the findings of \newcite{levy2014dependencywordembeddings} exploring the effectiveness of various dependency-based word embeddings on domain similarity, word similarity, and two other downstream tasks in English (question-type classification and named entity recognition.). They found that embeddings trained with UD contexts excel only in some tasks but not always improve the performance.}

It is our belief that fine-grained syntactic information such as the features we implemented in \S\ \ref{sec:features}, when added to an already robust architecture, could capture important structures of language and therefore boost the  performance of the system. We try to make this hypothesis evident by building the third and last experimental setting, as follows.

\subsection{Syntactically-informed BERT for Irony Detection}\label{sec:results-lang-models-autoencoders}
Lastly, we performed experiments with the state-of-the-art BERT language model. For each language, we ran the base cased M-BERT model as anticipated in Section \ref{subsec:models}.
In a second phase of this setting, we implemented the base architecture by adding the dependency-based syntactic features detailed in \ref{sec:features} in three different fashions in order to have a clear-cut evidence of the actual contribution derived from dependency syntax to irony detection.

\begin{table}[h]
\centering
\resizebox{\linewidth}{!}{
\begin{tabular}{lll|c|cccc}
\multirow{2}{*}{\textbf{language}} &
\multicolumn{2}{l|}{\multirow{2}{*}{\textbf{shared task winner (report and score)}}}
&
{\textbf{\textit{SVC}}} &
\multicolumn{4}{c}{\textbf{BERT}} \\
\cmidrule{5-8}
& & &
\textbf{\textit{+unigrams}} &
{\color{white}{---} ---\color{white}{---}} &
\textbf{+syntax} &
\textbf{+best\_feats} &
\textbf{+autoen.} \\
\toprule
English  & \cite{wu2018thu_ngnSemEval} & .705 &
\textit{.649} &
.655 &
.682 {\color{darkgreen} ($\uparrow$ .027)} &
.694 {\color{darkgreen} ($\uparrow$ .039)} &
.706 {\color{darkgreen} ($\uparrow$ .051)} \\
\midrule
Spanish  & \cite{gonzalez2019elirfirosva} & .683 &
\textit{.613} &
.663 &
.668 {\color{darkgreen} ($\uparrow$ .003)} &
.677 {\color{darkgreen} ($\uparrow$ .014)} &
.679 {\color{darkgreen} ($\uparrow$ .016)} \\
\midrule
French   & \cite{rouvier2017liadeft} & .783 &
\textit{.617} &
.770 & 
.785 {\color{darkgreen} ($\uparrow$ .015)} &
.772 {\color{darkgreen} ($\uparrow$ .002)} &
.679 {\color{red} ($\downarrow$ .091)} \\
\midrule
Italian  & \cite{cimino2018multitasklearning} & .731 &
\textit{.578} &
.699 &
.703 {\color{darkgreen} ($\uparrow$ .004)} &
.687 {\color{red} ($\downarrow$ .012)} &
.696 {\color{red} ($\downarrow$ .003)} \\
\bottomrule
\end{tabular}
}
\caption{Results obtained combining M-BERT and dependency-based syntactic features. Green values and arrows pointing up show an increment in performance, while red values and arrows pointing down indicate a performance reduction, with respect to results obtained by the bare architecture. }\label{tab:multilingual-bert-results}
\end{table}

\noindent In Table \ref{tab:multilingual-bert-results} we report the results of the best participating system in each one of the shared tasks (with the reference to their working notes). Furthermore, as a baseline reference measure, we also added the results obtained with a SVC and a bag of words of unigrams, as it was a baseline proposed in all competitions.
Each of the experiments with M-BERT has been performed 5 times with the hyper-parameters previously described in Section \ref{subsec:models} in order to take into account the differences of random initialization, and the average macro F1 score of such number of runs is reported.

It is interesting to see that, firstly, models implemented with the addition of dependency-based syntactic features obtain results in line with the state of the art in all four language scenarios (see shared tasks results). 
Moreover, in all of them, the addition of syntactic knowledge (\textit{M-BERT+syntax}) determined an improvement of scores with respect to the models where syntax is not taken into account (\textit{M-BERT}), as highlighted with the green values. 
Therefore, it seems that syntax plays an important role in the detection of irony and surely deserves further investigation exploring also more complex neural architectures.

We carried out a third set of experiments, which produced the scores reported in the table as \textit{M-BERT+best\_feats}. In these experiments we paired the M-BERT architecture with the best set of features extracted in the very first setting (see Section \ref{sec:results-classical-ML}).
What we observe in this case is that the extraction of the best features, and the subsequent reduction of the dimension of the feature space, is beneficial for English and Spanish. As a matter of fact, using the reduced feature set, we have an increment of the macro F1 score from $.682$ to $.694$ ($+ .012$) in English and from $.668$ to $.677$ ($+ .009$) in Spanish.  
On the other hand, French and Italian do not seem to profit from the reduction of available features.

\noindent Lastly, we tried a different technique for reducing the sparsity of features derived from syntax knowledge, implementing an \emph{autoencoder} \cite{bengio2009learning}, a well-known technique  to perform dimensionality reduction. The results of \textit{M-BERT+autoencoder} for all languages are comparable to those obtained with the other settings, and especially in English they are close to the state-of-the-art results and even surpass it by few points. For French and Italian, the autoencoding technique does not seem to be beneficial at all (see red values). It obtains, indeed, the worst results among all the performances.

All the above-mentioned conditions prove themselves interesting and surely lead us in the direction of further investigation, pointing mainly towards a better understanding of features' behavior when stacked in a pre-trained language model such as BERT. Furthermore, taking into account that irony is a pragmatic device inherently related to culture and language, our findings stress the importance of investigating features monolingually in order to provide a solid background for enhancing a multilingual system.


\subsection{Error Analysis}
From a lexical point of view, our qualitative analysis of results shows that even in the presence of clear lexical cues, like specific hashtags, classification errors occur in all datasets. This piece of evidence does not come unexpected. It is indeed a confirmation that our classification method --almost exclusively based on morpho-syntactic information-- which we developed in order to highlight the impact of syntax alone in the task of irony detection, is not especially influenced by the presence of such cues.

On the other hand, an error analysis based on the results generated from the best runs in the four languages also provides useful evidences about the impact of syntactic features to the detection of irony.
The distribution of Part-of-Speech tags and dependency relations in the whole test sets with respect to their distribution in the misclassified tweets is indeed unbalanced.
Namely, 
a higher number of \texttt{SYM}\footnote{See: \url{https://universaldependencies.org/u/pos/SYM.html}.}
and \texttt{X}\footnote{See: \url{https://universaldependencies.org/u/pos/X.html}.},
with respect to PoS-tags, and \texttt{parataxis}\footnote{See: \url{https://universaldependencies.org/u/dep/parataxis.html}.},
\texttt{flat}\footnote{See: \url{https://universaldependencies.org/u/dep/flat.html}.}
and \texttt{expl}\footnote{See: \url{https://universaldependencies.org/u/dep/expl.html}.},
with respect to dependency relations, characterizes tweets that are wrongly classified (either false positives or false negatives).
Regardless of the variation of the frequency of these morphological categories and syntactic relations in the four different languages, their incidence can be observed in the results. For instance, the presence of tokens tagged as \texttt{SYM} in the English misclassified tweets is higher by +22\% with regard to the average distribution, in Spanish is +7\%, in Italian is +4\%, while it does not seem to have particular relevance in the French data.

The categories and relations observed are typically found in user-generated texts and are moreover known as inherently hard to parse for the state-of-the-art UD parsers which, still nowadays, are mostly trained on standard texts and do not take into account idiosyncrasies of social media texts (emojis, hashtags, non-standard word forms). A debate regarding this one and other related issues is indeed ongoing in the UD community and a discussion can be seen, e.g., in a recent work by \newcite{sanguinetti2020treebanking}.

\section{Conclusion and Future Work}
In the present paper, we have investigated different scenarios regarding irony detection exploiting dependency-based syntactic features in a multilingual perspective.
Our current findings provide a meaningful support to the hypothesis that morpho-syntactic knowledge extracted from treebanks can be usefully exploited for addressing the irony detection task. In particular, they pave the way for a further investigation where the combination of a dependency-based syntactic approach and state-of-the-art neural models can be explored.

In the future, in order to validate our findings, we will propose a novel and wider experimental setting, where more languages are included, e.g. 
testing our approach on other languages for which both irony-annotated datasets and UD resources are available, 
such as Arabic \cite{ghanem2020irony,seddah2020arabizi} or German\footnote{\url{http://kti.tugraz.at/staff/rkern/courses/kddm2/2018/reports/team-27.pdf}.} \cite{rehbein2019tweede}.



\section*{Acknowledgments}
The work of C. Bosco and V. Basile was partially funded by Progetto di Ateneo/CSP 2016 Immigrants, Hate
and Prejudice in Social Media (S1618\_L2\_BOSC\_01). The work of M. Sanguinetti is funded by PRIN 2017 (2019-2022) project \textit{HOPE - High quality Open data Publishing and Enrichment}. The work of P. Rosso was partially funded by the Spanish MICINN under the research project MISMIS-FAKEnHATE on Misinformation and Miscommunication in social media: FAKE news and HATE speech(PGC2018-096212-B-C31) and by the Generalitat Valenciana under the project DeepPattern (PROMETEO/2019/121). Thanks also to P. Chiril for her suggestion on implementation methods.

\bibliography{bibliography}
\bibliographystyle{acl}

\end{document}